\documentclass[letterpaper, 10 pt, conference]{ieeeconf}  % Comment this line out if you need a4paper

\IEEEoverridecommandlockouts                              % This command is only needed if 
\usepackage{graphics} % for pdf, bitmapped graphics files
\usepackage{epsfig} % for postscript graphics files
\usepackage{amsmath} % assumes amsmath package installed
\usepackage{amssymb}  % assumes amsmath package installed
\usepackage{amsfonts}
\usepackage{cite}
\usepackage{url}
\usepackage{hyperref}
\usepackage[dvipsnames]{xcolor}
\usepackage{textcomp}
\usepackage{color}

\usepackage{times,amsmath,amsfonts,stmaryrd,multirow,subfigure,epsfig,enumerate}
\usepackage{breakurl}
\usepackage{flushend}
\usepackage{cite}
\usepackage{array}
\usepackage{pdfsync}
\usepackage{textcomp}
\usepackage{graphicx}
\usepackage{xfrac}
\usepackage{subfiles}
\usepackage{hyperref}

\hypersetup{colorlinks=true,citecolor=blue,linkcolor=blue,linktocpage=true}

\title{\LARGE \bf
Can We Use Diffusion Probabilistic Models for 3D Motion Prediction?
}

\author{Hyemin Ahn$^{1}$, Esteve Valls Mascaro$^{2}$, Dongheui Lee$^{2, 3}$% <-this % stops a space
\thanks{$^{1}$Hyemin Ahn is with Artificial Intelligence Graduate School (AIGS), Ulsan National Institute of Science and Technology (UNIST), Ulsan, Korea (e-mail: \texttt{hyemin.ahn@unist.ac.kr}).}%
\thanks{$^{2}$Esteve Valls Mascaro and Dongheui Lee are with Autonomous Systems, Technische Universität Wien (TU Wien), Vienna, Austria (e-mail: \texttt{\{esteve.valls.mascaro, dongheui.lee\}@tuwien.ac.at}).}%
\thanks{$^{3}$Dongheui Lee is also with the Institute of Robotics and Mechatronics (DLR), German Aerospace Center, Wessling, Germany.}%
}

\begin{document}

\maketitle
\thispagestyle{empty}
\pagestyle{empty}

%%%%%%%%%%%%%%%%%%%%%%%%%%%%%%%%%%%%%%%%%%%%%%%%%%%%%%%%%%%%%%%%%%%%%%%%%%%%%%%%
\begin{abstract}
After many researchers observed fruitfulness from the recent diffusion probabilistic model,
its effectiveness in image generation is actively studied these days. 
In this paper, our objective is to evaluate the potential of diffusion probabilistic models for 3D human motion-related tasks.
To this end, this paper presents a study of employing diffusion probabilistic models to predict future 3D human motion(s) from the previously observed motion.
% However, to the best of our knowledge, not many studies exist to evaluate the potential of diffusion probabilistic models for 3D motion-related tasks.
% To validate how much the diffusion probabilistic models can be effective for 3D motion-related tasks, this paper presents a study of employing diffusion probabilistic models to predict future 3D human motion(s) from the previously observed 3D motion.
Based on the Human 3.6M and HumanEva-I datasets, our results show that diffusion probabilistic models are competitive for both single (deterministic) and multiple (stochastic) 3D motion prediction tasks, after finishing a single training process.
In addition, we find out that diffusion probabilistic models can offer an attractive compromise, since they can strike the right balance between the likelihood and diversity of the predicted future motions.
Our code is publicly available on the project website: \textcolor{MidnightBlue}{\url{https://sites.google.com/view/diffusion-motion-prediction}}.
% However, we would not like to argue that diffusion models perform the best in  
% However, we would not like to argue that diffusion models can replace all kinds of existing state-of-the-art approaches for motion prediction tasks. 
% Rather, we would claim that diffusion models offer an attractive compromise that can balance the trade-off between the likelihood and diversity of the predicted future motions.
\end{abstract}

%%%%%%%%%%%%%%%%%%%%%%%%%%%%%%%%%%%%%%%%%%%%%%%%%%%%%%%%%%%%%%%%%%%%%%%%%%%%%%%%
\section{Introduction}

% deterministic vs generative 

Estimating how a human would move in the near future is an essential task for various applications such as surveillance \cite{surv_1, surv_2}, autonomous driving \cite{drive_1, drive_2}, and human-robot/computer-interaction \cite{hri_1}.
Many approaches have been proposed to solve this problem, often based on the motion capture datasets such as Human3.6M \cite{h36m_pami} or SMPL \cite{loper2015smpl}-based datasets such as AMASS \cite{mahmood2019amass}. 
In this paper, we concern with a task whose goal is to predict a sequence of 3D pose skeletons in Human3.6M and HumanEva-I \cite{sigal2006humaneva} datasets, when a previously observed 3D pose sequence is given as an input.

Existing works on 3D skeleton motion prediction can be categorized as follows. 
One line of research focuses on models for deterministic motion prediction \cite{deter_1, deter_2, s-rnn, DCT-GCN, ST-TR, 2CH-TR}. 
These works aim at predicting a single motion that is most likely to be observed in the future. 
Therefore, their performance is usually evaluated based on an $L2$-distance between a prediction and a ground truth.
% measured from the 3D pose skeletons in euler-angle representation.
Another line of research focuses on generative models for stochastic motion prediction \cite{hp-gan, mt-vae, Dlow, gsps}. 
Their performance is evaluated based on the metrics for likelihood and diversity.
After generating a fixed number of prediction samples from a single observation,
the likelihood is measured based on the minimum distance between the prediction samples and ground truth, and the diversity is measured based on the average distance between all pairs of prediction samples.

\begin{figure}
    \centering
    \includegraphics[width=0.45\textwidth]{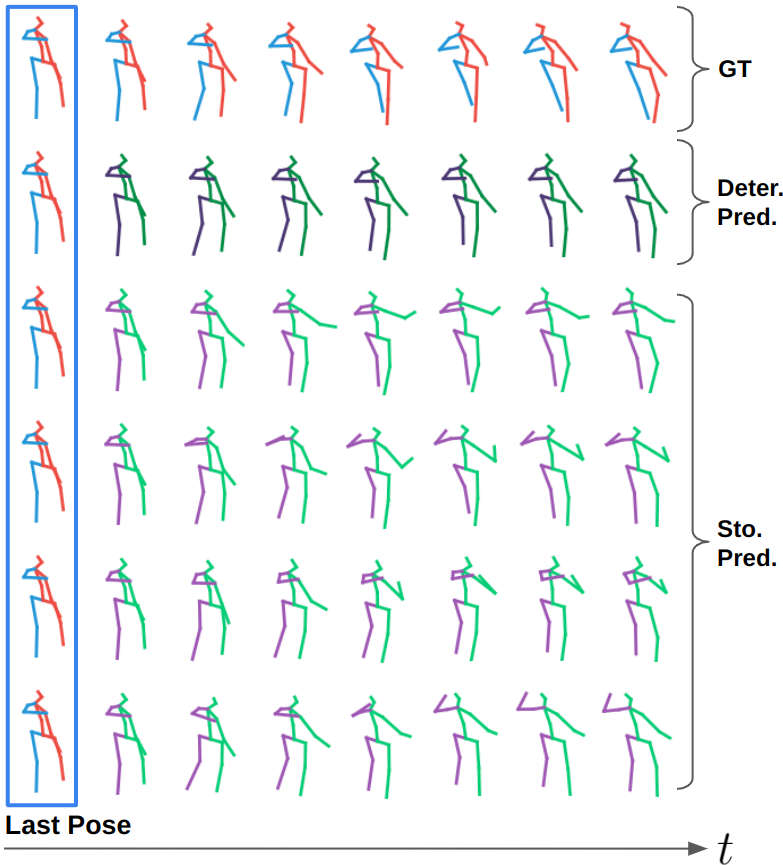}
    \caption{Example results when diffusion probabilistic models are used for 3D human motion prediction tasks, when observed motion is `walking'. After a single training procedure, diffusion models can be effectively used for both deterministic (Deter.) and stochastic (Sto.) motion prediction tasks.}
    \label{fig:first}
    \vspace{-3mm}
\end{figure}

However, we cannot judge which approach is always better than the other, since the efficiency would depend on the target application values.
For instance, when one needs only the most precise sample with low latency, deterministic approaches would be better. 
If we say that both approaches are necessary, our next question would be whether we can propose an efficient model for both types of prediction.
To answer the question, we study the possibility of using diffusion probabilistic models \cite{ddpm, csdi} for both deterministic and stochastic 3D motion prediction tasks.

If we propose a diffusion probabilistic model \cite{ddpm, csdi} as a solution, one might ask us whether this is because we are fascinated by its performance in image generation \cite{imagen, palette}.
Frankly speaking, yes, we initiated this study out of our curiosity -- can we use diffusion probabilistic models for 3D motion prediction?
Unfortunately, our experimental results show that the diffusion model cannot perfectly replace existing state-of-the-arts for both deterministic and stochastic motion prediction tasks.
However, we found a glimpse of hope in diffusion models, due to their effectiveness in both prediction types after a single training procedure, and their ability to properly balance the trade-off between diversity and likelihood.

Figure~\ref{fig:first} shows the example results when the diffusion models are used for both deterministic and generative motion prediction tasks.
Although a diffusion model is essentially a generative model, we found that the deterministic sample with a fair performance can be obtained from the diffusion model when all randomness is excluded from its denoising process. 
% mention the comparison result with VAE and GAN
In addition, we found out that the diffusion models can fix the flaws of several generative methods \cite{Dlow}, which highlight the diversity of generated samples. Existing works as \cite{Dlow} claim that the likelihood of predicted samples is high when the minimum distance between samples and ground truth is low. Because of this, \cite{Dlow} can often generate the motions that are out-of-context as \cite{contexuallyplausible} pointed out. Compared to this, our diffusion models can generate prediction samples that are more likely to occur, so the generated motion does not diverge too much to be called out-of-context.

The remaining paper is constructed as follows.
After representing our literature survey in Section~\ref{sec:rel},
Section~\ref{sec:met} will explain how general diffusion models work as well as how we design ours to solve 3D motion prediction tasks.
Section~\ref{sec:exp} will show both qualitative and quantitative experiment results, and a related discussion will be also presented. Finally, this paper will end in Section~\ref{sec:con} by mentioning limitations and future works. 

% The contributions of this paper can be summarized as follows:
% \begin{itemize}
%     \item We the potential of diffuse probabilistic models for the deterministic or stochastic 3D human motion prediction tasks. 
%     \item
% \end{itemize}

\section{Related Work} \label{sec:rel}

\subsection{3D Motion Prediction}
\noindent \textbf{Deterministic Models.}
The goal of deterministic 3D motion prediction is to minimize the distance between a predicted motion and ground truth. 
To solve this problem, early works relying on deep neural networks \cite{s-rnn, deter_1, deter_2} often employed recurrent neural networks (RNNs) \cite{lstm, gru}, which are still well-known for their effectiveness in processing time-series data. 
Among RNN-based works, a notable model is a structure RNN (S-RNN) \cite{s-rnn}, which considers the spatio-temporal information of human motion, by manually designing the high-level spatio-temporal graph to explicitly model the human body structure (i.e., spine, arm, and leg).

While S-RNN understands the human body structure based on the handcrafted network structure, there is another line of research \cite{DCT-GCN, mao2020history} that uses graph convolutional network (GCN) to overcome this manually designed spatial relationship understanding. For instance, \cite{DCT-GCN} suggested a model named DCT-GCN, where discrete cosine transform (DCT) understands the temporal information of motion, and GCN learns the spatial relationship between human body joints. DCT-GCN obtains the state-of-the-art result when evaluated on Euler-angle-based mean squared error, but its best result can be obtained when the model is separately trained for each short- or long-term prediction.

Recently, several works for deterministic motion prediction \cite{ST-TR, 2CH-TR} are based on the Transformer \cite{transformer}, which was originally suggested for language understanding problems.
Models named Spatio-Temporal Transformer (ST-TR) \cite{ST-TR} and 2-Channel Transformer (2CH-TR) \cite{2CH-TR}, understand the spatio-temporal relationship of human motion by putting the self-attention mechanism on each pose-parameter (spatial) and time (temporal) dimension.
After understanding each spatial and temporal information in parallel, outputs from both attention mechanisms are properly combined. The difference between ST-TR and 2CH-TR comes from when and how often the model combines spatial and temporal information.

\vspace{2mm}
\noindent \textbf{Generative Models.} 
The goal of stochastic 3D motion prediction is to build a generative model which can sample out several future motions that are likely to happen after the observed human motion.
To solve this problem, early works \cite{hp-gan, mt-vae} employed deep generative models such as variational autoencoders (VAEs) \cite{vae} or generative adversarial networks (GANs) \cite{gan}.
For instance, \cite{mt-vae} suggested a generative model based on the conditional VAEs, and showed that VAEs can sample out several future motions that are reasonable as well as diverse.
Compared to VAE, \cite{hp-gan} showed that GANs based on the Wasserstein loss function can be effectively used in stochastic motion prediction tasks.

While these works \cite{hp-gan, mt-vae} focused on exploring the potential of using deep generative models in stochastic motion prediction tasks, another line of works \cite{Dlow, gsps} focused on sampling out as much as diverse motions that can contain the most plausible motion at the same time.
For instance, \cite{Dlow} proposed to train a post-hoc model which can be attached to the pre-trained deep generative model. This post-hoc model maps a random variable to several latent vectors of the pre-trained generative model.
Based on the \textit{diversity-promoting prior}, the post-hoc model is trained to improve the diversity between samples, which can be obtained by decoding the mapped latent vectors.

Experiments in \cite{Dlow, gsps} evaluate the likelihood of prediction samples based on the \textit{minimum} distance between the samples and the ground truth(s).
They denote the prediction samples as plausible based on the sample that is closest to the ground truth(s).
However, this can make it difficult for users to choose the most plausible motion among the prediction samples, since all samples will not be distributed near the most plausible motion.
For instance, if the observed motion is a human sitting down and drinking something, \cite{Dlow} and \cite{gsps} can produce motion samples that predict the human suddenly standing up and starting discussing something with others. 
As \cite{contexuallyplausible} has pointed out, we would like to also focus on the necessity of contextually plausible and diverse motion sampling. Therefore, our paper would evaluate the likelihood of prediction also based on the mean and standard deviation of distances between the samples and ground truth.

\subsection{Diffusion Probabilistic Models}
% add text to motion task based on the diffusion model.
Diffusion probabilistic models \cite{ddpm} have become a new rising star in generative models after showing excellent performance in image synthesis. Especially, its performance on text-conditioned image synthesis \cite{imagen} makes researchers as well as the public in awe. 
Diffusion models consider two processes: a forward process that slowly destructs the data sample by gradually injecting the random noise, and a reverse process that learns how to reconstruct the data sample by gradually denoising the random noise.
While the advantage of diffusion models can be empirically shown based on their performances, the disadvantage is the speed of their sampling process. If the reverse process includes $1000$ times of denoising processes, it means that the data sample can be obtained after feed-forwarding the random noise to the denoising network for $1000$ times. Of course, this disadvantage can be circumvented if the application does not require the prediction samples with low latency.

Aside from image generation tasks, nowadays researchers are suggesting to use diffusion models in various generation tasks, such as text-to-speech \cite{diff_tts}, text-to-sound \cite{diff_sound}, and video \cite{ho2022video}.
Focusing on motion-related tasks like ours, several works incorporate diffusion models in text-conditioned motion generation tasks \cite{kim2022flame, zhang2022motiondiffuse}. 
For the motion of intelligence agents, \cite{janner2022planning} suggests using diffusion models to sample out trajectories for properly solving a given task.
In our paper, we use diffusion models in 3D human motion prediction tasks, but to the best of our knowledge, there is no attempt yet to use diffusion models in the 3D motion prediction task. 
But we believe more researchers would involve in using diffusion models to answer this question -- can diffusion models be our new savior in any kind of data generation tasks?

\section{Method} \label{sec:met}
\subsection{Preliminaries}
We will provide a short description of diffusion probabilistic models first. Note that our description relies on \cite{ddpm} and \cite{csdi}, which provide a basis for our work.
\hfill \break \newline
\textbf{Diffusion Probabilistic Model.}
Let $\mathbf{x}^0 \sim q(\mathbf{x}^0)$ denote a data point sampled from its distribution $q$. In order to learn $p_\theta(\mathbf{x}^0)$ which can model $q(\mathbf{x}^0)$, diffusion probabilistic models consider two processes. 
One is a \textit{forward process} which gradually deconstructs $\mathbf{x}^0$ by injecting a subtle Gaussian noise for $K$ times, such that $\mathbf{x}^0$ can be destroyed into $\mathbf{x}^1,\ldots,\mathbf{x}^K$,  where $p(\mathbf{x}^K)=\mathcal{N}(\mathbf{0}, \mathbf{I})$. 
This process can be formulated as below, which is to follow a Markov chain $q(\mathbf{x}^k|\mathbf{x}^{k-1})$ for $K$ times:
\begin{align}
    q(\mathbf{x}^{1:K}|\mathbf{x}^0)&=\prod_{k=1}^K q(\mathbf{x}^k|\mathbf{x}^{k-1}) \\ q(\mathbf{x}^k|\mathbf{x}^{k-1})&=\mathcal{N}(\sqrt{1-\beta_k}\mathbf{x}^{k-1}, \beta_k\mathbf{I}),
\end{align}
where $\beta_k$ denotes a constant for a noise level. Note that $\mathbf{x}^k$ can be sampled from $\mathbf{x}^0$ directly with a closed-form solution:
\begin{equation}
    \mathbf{x}^k = \sqrt{\alpha_k}\mathbf{x}^0 + \sqrt{1-\alpha_k}\boldsymbol{\epsilon}, \;\; \boldsymbol{\epsilon}\sim \mathcal{N}(\mathbf{0, I}),
\end{equation}
where $\hat{\alpha}_k=1-\beta_k$ and $\alpha_k=\prod_{i=1}^k \hat{\alpha}_i$.

\begin{figure*}
    \centering
    \includegraphics[width=0.975\textwidth]{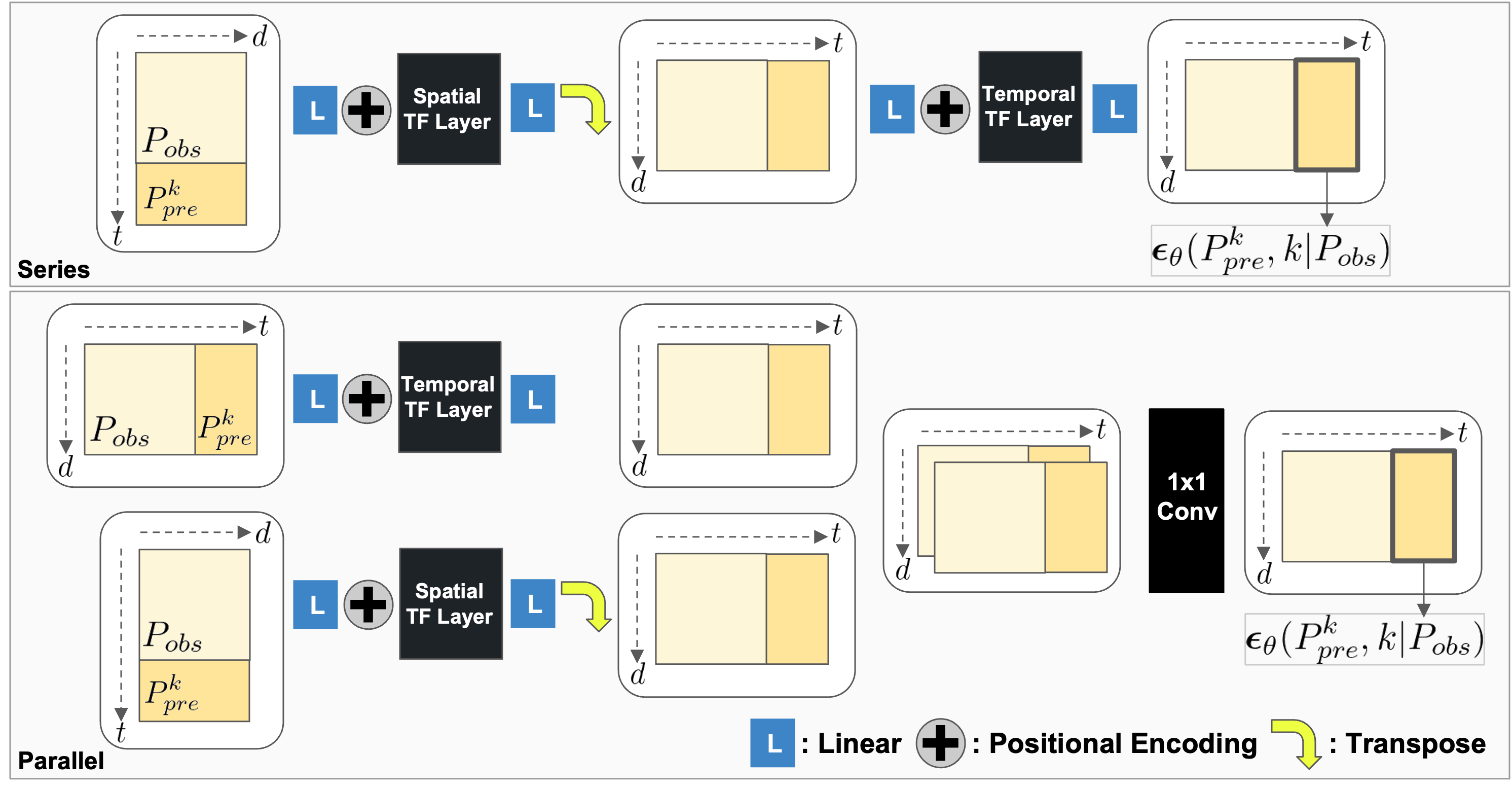}
    \caption{Two designs of our Transformer-based motion denoiser. Inspired by ST-TR \cite{ST-TR}, 2CH-TR \cite{2CH-TR} and CSDI \cite{csdi}, our motion denoiser processes both spatial and temporal information in series (top) or in parallel (bottom). Here, $d$ and $t$ stand for the dimension of each pose-parameter and time, and TF stands for Transformer \cite{transformer}. Note that the positional encoding also involves adding a learnable vector that represents a diffusion step $k$ as \cite{csdi} suggests.}
    \label{fig:net}
\end{figure*}

Another is a \textit{reverse process}, which goal is to obtain $\mathbf{x}^0$ starting from $\mathbf{x}^K \sim \mathcal{N}(\mathbf{0, I})$, by gradually denoising $\mathbf{x}^K$. 
This process can also be formulated as following a Markov chain $p_\theta(\mathbf{x}^{k-1}|\mathbf{x}^k)$ for $K$ times:

\begin{align}
    p_\theta(\mathbf{x}^{0:K}) &= p(\mathbf{x}^K) \prod_{k=1}^K p_\theta(\mathbf{x}^{k-1}|\mathbf{x}^k), \\
    p_\theta (\mathbf{x}^{k-1}|\mathbf{x}^k) &= \mathcal{N} \big( \mathbf{x}_{t-1}; \boldsymbol{\mu}_\theta(\mathbf{x}^k, k), \sigma^2(k)\mathbf{I} \big),
\end{align}
where $p(\mathbf{x}^K)=\mathcal{N}(\mathbf{0}, \mathbf{I})$.
To obtain $\boldsymbol{\mu}_\theta$ and $\sigma$, \cite{ddpm} suggests denoising diffusion probabilistic models (DDPM), which get $\sigma^2(k) = \frac{1-\alpha_{k-1}}{1-\alpha_{k}}\beta_k$, parameterize $\boldsymbol{\mu}_\theta$ with $\theta$, and sample $\mathbf{x}^{k-1}\sim p_{\theta}(\mathbf{x}^{k-1}|\mathbf{x}^k)$ as below:
\begin{align}
    \boldsymbol{\mu}_\theta(\mathbf{x}^k, k) &= \frac{1}{\sqrt{\hat{\alpha}_k}} \bigg( \mathbf{x}^k - \frac{\beta_k}{\sqrt{1-\alpha_k}} \boldsymbol{\epsilon}_\theta(\mathbf{x}^k, k) \bigg). \\
    \mathbf{x}^{k-1} &= \boldsymbol{\mu}_\theta(\mathbf{x}^k, k) + \sigma(k)\mathbf{z}, \;\;\mathbf{z} \sim \mathcal{N}(\mathbf{0, I}).
\end{align}

In practice, $\boldsymbol{\epsilon}_\theta$ is modeled with a neural network, and it learns how much to denoise from $\mathbf{x}^k$. To train this, \cite{ddpm} suggested a simplified loss function as below:
\begin{align}
    \mathcal{L}(\theta) &= \| \boldsymbol{\epsilon} - \boldsymbol{\epsilon}_\theta(\mathbf{x}^k, k)\|^2 \nonumber \\
    &= \|\boldsymbol{\epsilon} - \boldsymbol{\epsilon}_\theta(\sqrt{\alpha_t}\mathbf{x}^0 + \sqrt{1-\alpha_t}\boldsymbol{\epsilon}, k)\|^2.
\end{align}

In a training process, $k$ is randomly sampled to obtain $\mathcal{L}(\theta)$. For more details, please refer to \cite{ddpm} and \cite{csdi}.
\vspace{2mm}

% Note that the forward process does not require learnable parameters. What diffusion models learn is how to model a \textit{reverse process}.

\noindent \textbf{Conditional Diffusion Model.}
A conditional score-based diffusion model for imputation (CSDI) \cite{csdi} is proposed to solve a time-series imputation problem using diffusion models. 
It adds conditional information $\mathbf{x}_{co}$ to eq. (4)-(5):
\begin{align}
    p_{\theta}(\mathbf{x}^{0:K}) &= p(\mathbf{x}^K) \prod_{k=1}^K p_\theta(\mathbf{x}^{k-1}|\mathbf{x}^{k}, \mathbf{x}_{co}),\\
    p_\theta(\mathbf{x}^{k-1}|\mathbf{x}^{k}, \mathbf{x}_{co}) &= \mathcal{N}(\mathbf{x}^{k-1}; \boldsymbol{\mu}_\theta(\mathbf{x}^{k}, k | \mathbf{x}_{co}), \sigma^2(k)\mathbf{I})
\end{align}

To define $\boldsymbol{\mu}_\theta(\mathbf{x}^{k}, k | \mathbf{x}_{co})$,
eq. (6)-(7) can be rewritten by adding $\mathbf{x}_{co}$ as a condition to $\boldsymbol{\mu}_\theta$ and $\boldsymbol{\epsilon}_\theta$.
Note that $\boldsymbol{\epsilon}_\theta(\mathbf{x}^{k}, k | \mathbf{x}_{co})$ is modeled with a neural network to learn how much to denoise from $\mathbf{x}^k$ given $\mathbf{x}_{co}$.
When training the network, the same loss function as eq. (8) is used, by replacing $\boldsymbol{\epsilon}_\theta$ properly with $\mathbf{x}_{co}$ as a condition.

\subsection{Problem Formulation}
Let $\mathbf{p}_t \in \mathbb{R}^{D}$ be a 3D pose vector at time $t$, which can be denoted with various representations such as axis-angle, Euler-angle, or $xyz$-position. Here, $D=3n$ and $n$ denotes the number of joints. 
A task of 3D human motion prediction can be defined as predicting future $L$ poses,
${P}_{pre}=\{\mathbf{p}_{T+1},\ldots \mathbf{p}_{T+L}\} \in \mathbb{R}^{L \times D}$,
when $T$ poses, ${P}_{obs}=\{\mathbf{p}_1,\ldots \mathbf{p}_{T}\} \in \mathbb{R}^{T \times D}$ are observed.
%

% If one designs a deterministic model [cite] for this 3D motion prediction, the main goal would be finding a prediction model $f_{\theta_d}(P_{obs})$, which can minimize below loss function:
% \begin{equation}
%     \mathcal{L}(\theta_d) = \| {P}_{pre} - \hat{P}_{pre} \|_2,\;\;\hat{P}_{pre}=f_{\theta_d}({P}_{obs}),
% \end{equation}
% where $\theta_d$ denotes the parameters of the deterministic model.

% On the other hand, if one designs a generative model [cite] for this 3D motion prediction, the main goal would be learning a distribution $p_{\theta_g}(P_{pre}|P_{obs})$, which can approximate a true data distribution $q(P_{pre}|P_{obs})$. Here, $\theta_g$ denotes the parameters of the generative model. After training, we can sample out possible $P_{pre}$'s when $P_{obs}$ is given, where this sampling process in practice can be denoted as below:
% \begin{equation}
%     \hat{P}_{pre} = f_{\theta_g}(P_{obs}, \mathbf{z}),\;\;\mathbf{z} \sim \mathcal{N}(\mathbf{0}, \mathbf{I}). 
% \end{equation}

We utilize CSDI \cite{csdi} for obtaining $P_{pre}$ from given $P_{obs}$.
Starting from $P_{pre}^0=P_{pre}$, our forward process can obtain $P_{pre}^k$ as below:
\begin{equation}
    P_{pre}^k = \sqrt{\alpha_k} P_{pre}^0 + \sqrt{1-\alpha_k} \boldsymbol{\epsilon}, \;\; \boldsymbol\epsilon \sim \mathcal{N}(\mathbf{0, I})
\end{equation}
For a reverse process, we propose a denoiser network which models $\boldsymbol{\epsilon}_\theta(\mathbf{x}^k,k| \mathbf{x}_{co})=\boldsymbol{\epsilon}_\theta(P_{pre}^k, k | P_{obs})$.
This network is trained by minimizing
$\mathcal{L}(\theta) = \| \boldsymbol{\epsilon} - \boldsymbol{\epsilon}_\theta(P_{pre}^k, k| P_{obs}) \|^2$.

After training, we can sample $P_{pre}^0$ by repeating below reverse process for $K$ times, starting from $P_{pre}^K \sim \mathcal{N}(\mathbf{0, I})$:
\begin{equation}
    P_{pre}^{k-1} = \boldsymbol{\mu}_\theta(P_{pre}^k, k | P_{obs}) + \sigma(k)\mathbf{z}, \;\; \mathbf{z}\sim\mathcal{N}(\mathbf{0, I}),
\end{equation}%
where $\boldsymbol{\mu}_\theta(P_{pre}^k, k | P_{obs})$ is defined with $\boldsymbol{\epsilon}_\theta(P_{pre}^k, k | P_{obs})$ and properly modified version of eq. (6).
After finishing training, if our denoiser network is used for deterministic prediction in the test phase, we set $P_{pre}^K$ and $\mathbf{z}$ as zero-vectors, such that all randomness in eq. (12) can be ignored.

\subsection{Transformer-based Motion Denoiser}
Since $P_{pre}^k$ and $P_{obs}$ are time-series of human pose vectors, one can model
$\boldsymbol{\epsilon}_\theta(P_{pre}^k, k | P_{obs})$ with neural network architectures which can understand time-series data.
For example, network architectures such as RNNs \cite{lstm, gru} or Transformers \cite{transformer} can be candidates.
We empirically found out that the denoisers based on the Transformers that process both spatial and temporal information are most effective.

Figure~\ref{fig:net} shows how we design our Transformer-based denoisers in two ways.
Inspired by \cite{csdi}, the first denoiser shown on top of the figure processes both information in series. After concatenating $P_{pre}^k \in \mathbb{R}^{L \times D}$ and $P_{obs} \in \mathbb{R}^{T \times D}$ such that input can be $P_{inp}^k \in \mathbb{R}^{(T+L) \times D}$, $P_{inp}^k$ passes spatial and temporal transformer layers in series, where each layer applies self-attention to time and pose-parameter dimension.
Before passing each transformer layer, positional encoding is added to the input as \cite{transformer} suggests, with respect to pose-parameter $d\in[0, D]$ (spatial) or time $t\in[0, T]$ (temporal) dimension.
Also, the additional learnable positional encoding that projects a diffusion step $k$ into a vector space is added to the input as \cite{csdi} suggests.
Let $P_{out}^k\in \mathbb{R}^{(T+L) \times D}$ denote the output which can be obtained after $P_{inp}^k$ passing two layers.
Then, the last $L \times D$ parts from $P_{out}^k$ is obtained as $\boldsymbol\epsilon_\theta(P_{pre}^k, k|P_{obs})$, which would be used for denoising $P_{pre}^k$.

The second denoiser shown on the bottom of Figure~\ref{fig:net}, is inspired by \cite{ST-TR} and \cite{2CH-TR}, and works in parallel to understand spatio-temporal information. 
After $P_{inp}^k$ passes both spatial and temporal transformer layers in parallel, two matrices with the same size as $P_{inp}^k$ are obtained, and concatenated into a 3rd-order tensor whose size is $2\times (T+L) \times D$. After this tensor passes 2-dimensional convolutional layer with $(1\times 1)$-sized kernel, the output $P_{out}^k\in \mathbb{R}^{(T+L) \times D}$ is obtained.
From $P_{out}^k$, $\boldsymbol\epsilon_\theta(P_{pre}^k, k|P_{obs})$ is obtained as same as in the first denoiser. 

Note that we do not use encoder-decoder based structure, which encode a set of feature vectors from $P_{obs}$ and decode $\boldsymbol\epsilon_\theta(P_{pre}^k, k|P_{obs})$ from the encoded feature vectors and $P_{pre}^k$.
We tried various denoisers of Transformer- or RNN-based encoder-decoder, but none of them turns out to be effective.

\subsection{Implementation Details}
Our transformer-based motion denoisers have a self-attention module with 8 multi-heads and 512-dimensional query, key, and value vectors. And each temporal or spatial transformer layer shown in Figure~\ref{fig:net} consists of a single-layered transformer encoder.
To train denoisers, we set batch size as 512 and update parameters for 50,000 iterations with Adam optimizer of learning rate $0.0001$.
The diffusion step is set as $k\in[0, 20]$, with linearly scheduled noise levels $\beta_k$ that ranges between $0.001$ ($k\downarrow$) and $0.333$ ($k\uparrow$). 

\section{Experiment} \label{sec:exp}
\subsection{Dataset and Metric}
\noindent \textbf{Dataset.}
We conduct our experiment for both deterministic and stochastic motion prediction tasks. For deterministic experiments, we use the Human3.6M dataset \cite{h36m_pami} and measure the Euler-angle mean square error (MSE) for evaluation as other works \cite{s-rnn, DCT-GCN, ST-TR, 2CH-TR} do. Here, with 25 fps, input observation has 50 frames, and output prediction has 25 frames.
For stochastic experiments, we preprocess Human3.6M \cite{h36m_pami} and HumanEva-I \cite{sigal2006humaneva} datasets into $xyz$-based representation as \cite{Dlow, gsps} do. Based on that, various metrics for evaluating likelihood and diversity are measured. Here, with 50 fps, an input observation has 25 frames, output prediction has 100 frames, and the number of prediction samples is 50.

\vspace{2mm}

\noindent \textbf{Metrics.}
As mentioned above, we measure the performance of our denoiser based on the Euler-angle MSE when it is used for deterministic prediction.
For stochastic prediction, we use several metrics from what \cite{Dlow} suggests to evaluate likelihood and diversity. But we propose more metrics such as aDE, sDE, aFDE, and sFDE to measure how the samples are distributed near the ground truth. Note that some of the below sentences describing metrics are borrowed from \cite{Dlow}.

(1) \textbf{Average Pairwise Distance (APD)}: average $L2$ distance between pairs from $N$ predictions $\hat{\mathbf{x}} \in \mathbb{R}^{L \times D}$, which is computed as $\frac{1}{N(N-1)}\sum_{i=1}^N \sum_{j\neq i}^N \| \hat{\mathbf{x}}_i - \hat{\mathbf{x}}_j \|_2$. This measures the diversity within $N$ predictions.
(2) \textbf{minimum Displacement Error (mDE)}: the minimum $L2$ distance between all $N$ predictions $\hat{\mathbf{x}}$ and ground truth $\mathbf{x}$, which is computed as $ \min_{\hat{\mathbf{x}}}\frac{1}{L}\|\hat{\mathbf{x}}- \mathbf{x} \|_2$.
This metric was defined as ADE in \cite{Dlow}.
(3) \textbf{average Displacement Error (aDE)}: the average $L2$ distance between all $N$ predictions $\hat{\mathbf{x}}$ and ground truth $\mathbf{x}$, which is computed as $\frac{1}{NL} \sum_{i=1}^N\|\hat{\mathbf{x}}_i- \mathbf{x} \|_2$.
(4) \textbf{standard deviation of Displacement Error (sDE)}: the standard deviation of $L2$ distances between all $N$ predictions and ground truth.
(5) \textbf{minimum Final Displacement Error (mFDE)}: the minimum $L2$ distance between final poses of $N$ predictions and ground truth, which is calculated as $\min_{\hat{\mathbf{x}}}\|\hat{\mathbf{x}}(L) - \mathbf{x}(L)\|_2$. This metric was defined as FDE in \cite{Dlow}.
(6) \textbf{average Final Displacement Error (aFDE)}: the average $L2$ distance between final poses of $N$ predictions and ground truth, which is calculated as $\frac{1}{N}\sum_{i=1}^N\|\hat{\mathbf{x}}_i(L) - \mathbf{x}(L)\|_2$.
(7) \textbf{standard deviation of Final Displacement Error (sFDE)}: the standard deviation of $L2$ distances between final poses of $N$ predictions and ground truth.

\subsection{Quantitative Results}
\noindent \textbf{Deterministic Prediction.} Table~\ref{tab:mse} compares Euler-angle MSEs when our diffusion model is used for deterministic motion prediction.
Here, bold fonts denote the best results among all approaches, and underlines denote the best results among our denoisers (series or parallel).
It is shown that the overall performance of DCT-GCN \cite{DCT-GCN} is still the best.
Among our approaches, the denoiser which understands spatial and temporal information in series is better than the parallel denoiser. 
Although our models do not achieve state-of-the-art results, it is shown that our approaches are better in long-term prediction (1000ms) when compared with other transformer-based models \cite{ST-TR, 2CH-TR}.
This is a notable result, since (1) our models are originally generative ones, and (2) our models do not require additional training for deterministic prediction since ignoring all randomness in the denoising process is all they need.

\begin{table}
\centering
\caption{Average MSE errors of Deterministic Motion Prediction}
\setlength\tabcolsep{5.5pt}
\begin{tabular}{l|llllll}
\textbf{milisecond (ms)}       & \multicolumn{1}{c}{\textbf{80}} & \multicolumn{1}{c}{\textbf{160}} & \multicolumn{1}{c}{\textbf{320}} & \multicolumn{1}{c}{\textbf{400}} & \multicolumn{1}{c}{\textbf{560}} & \multicolumn{1}{c}{\textbf{1000}}  \\ 
\hline
S-RNN \cite{s-rnn}                          & 0.933                           & 1.166                            & 1.397                            & 1.526                            & 1.711                            & 2.139                              \\
DCT-GCN \cite{DCT-GCN}                       & {0.295}                           & \textbf{0.542}                            & \textbf{0.857}                            & \textbf{0.974}                            & \textbf{1.154}                            & \textbf{1.590}                              \\
ST-TR \cite{ST-TR}          & 0.303                           & 0.550                            & 0.901                            & 1.021                            & 1.229                            & 1.722                              \\
2CH-TR \cite{2CH-TR}                  & \textbf{0.293}                           & 0.555                            & 0.893                            & 1.016                            & 1.245                            & 1.744                              \\ 
\hline
\textbf{Ours (Series)} &\underline{0.325}                           & \underline{0.615}                            & \underline{0.990}                            & \underline{1.128}                            & \underline{1.309}                            & 1.721  
\\
\textbf{Ours (Parallel)} & 0.350                           & 0.646                            & 1.007                            & 1.148                            & 1.317                            & \underline{1.688}                                    
\end{tabular}
\vspace{-17mm}
\label{tab:mse}
\end{table}

\begin{table*}
\centering
\caption{Diversity and Likelihood Metrics of Stochastic Motion Prediction}
\setlength\tabcolsep{4pt}
\begin{tabular}{l|lllllll|lllllll}
                         & \multicolumn{7}{c|}{Human 3.6M \cite{h36m_pami}}                                                                                                                             & \multicolumn{7}{c}{HumanEva-I \cite{sigal2006humaneva}}                                                                                        \\ 
\cline{2-15}
metrics                  & \multicolumn{1}{c}{APD$\uparrow$} & \multicolumn{1}{c}{mDE$\downarrow$} & \multicolumn{1}{c}{aDE$\downarrow$} & \multicolumn{1}{c}{sDE$\downarrow$} & mFDE$\downarrow$           & aFDE$\downarrow$           & sFDE$\downarrow$           & APD$\uparrow$            & mDE$\downarrow$           & aDE$\downarrow$           & sDE$\downarrow$           & mFDE$\downarrow$           & aFDE$\downarrow$           & sFDE$\downarrow$            \\ 
\hline
DLow \cite{Dlow}                    & \underline{11.741}          & \underline{0.425}            & 0.968                    & 0.355                    & \underline{0.518}  & 1.387          & 0.541          & \underline{4.855}  & 0.251          & 0.585          & 0.208          & 0.268          & 0.710          & 0.255           \\
VAEs \cite{vae, Dlow}                     & 6.852                   & 0.460                    & \underline{0.720}            & \underline{0.139}            & 0.557          & \underline{1.025}  & 0.243          & 2.299          & 0.265          & 0.426          & 0.083          & 0.299          & 0.562          & 0.137           \\
GSPS \cite{gsps}                    & \textbf{14.757}         & \textbf{0.389}           & 1.206                    & 0.623                    & \textbf{0.496} & 1.554          & 0.729          & \textbf{5.825} & \textbf{0.233} & 0.655          & 0.206          & \underline{0.244}  & 0.763          & 0.268           \\ 
\hline
\textbf{Ours (Series)}   & 7.587                   & 0.527                    & 0.764                    & \textbf{0.132}           & 0.669          & 1.093          & \textbf{0.228} & 2.746          & 0.257          & \underline{0.383}  & \underline{0.065}  & 0.260          & \underline{0.490}  & \underline{0.130}  \\
\textbf{Ours (Parallel)} & 6.445                   & 0.477                    & \textbf{0.719}           & \underline{0.139}            & 0.584          & \textbf{1.018} & \underline{0.234}  & 1.508          & \underline{0.242}  & \textbf{0.312} & \textbf{0.037} & \textbf{0.238} & \textbf{0.385} & \textbf{0.078}
\end{tabular}
\label{tab:sto}
\vspace{-3mm}
\end{table*}

\noindent \textbf{Stochastic Prediction}
Table~\ref{tab:sto} shows the comparison of metrics for measuring the likelihood and diversity. 
Here, bold fonts denote the best result and underlines denote the second best result among all approaches. 
It is shown that previous works \cite{Dlow, gsps} focusing on sample diversity best perform in APD.
Also, it is shown that they are generally better in terms of mDE and mFDE.
We would like to argue here that the high diversity in prediction increases the probability of having one sample closest to the ground truth.
Then, how can we choose the most plausible result among predictions that are sampled to be diverse?

This is the same question that \cite{contexuallyplausible} also pointed out. 
So in \cite{contexuallyplausible}, metrics for measuring the quality and context are proposed.
For measuring the quality, \cite{contexuallyplausible} used a pre-trained binary classifier which can discriminate the ground truths (real) from predictions (fake). If this classifier fails to discriminate the predicted motions as fake, a higher quality score is obtained.
For measuring the context, \cite{contexuallyplausible} used a pre-trained model which classifies action from motion. If it estimates that the action label of prediction is as same as the observed motion, a higher context score is obtained.

However, we were not able to use the same metric as \cite{contexuallyplausible} since its pre-trained classifiers were not openly released. 
Therefore, we instead propose metrics such as aDE, sDE, aFDE, and sFDE, to measure how closely the samples are distributed near the ground truth.
Results show that our approaches generally perform better in terms of these new metrics, and the parallel denoiser performs better than the series one. 
% But there could be a question of how other non-diffusion generative models would work. 
We also present the result from VAEs \cite{vae} that were implemented by \cite{Dlow}, to check how other non-diffusion generative models work. 
It is shown that the overall performances of our series/parallel denoiser in diversity and likelihood are generally better than the VAEs, especially in the HumanEva-I dataset. 

\begin{figure*}
    \centering
    \includegraphics[width=0.95\textwidth]{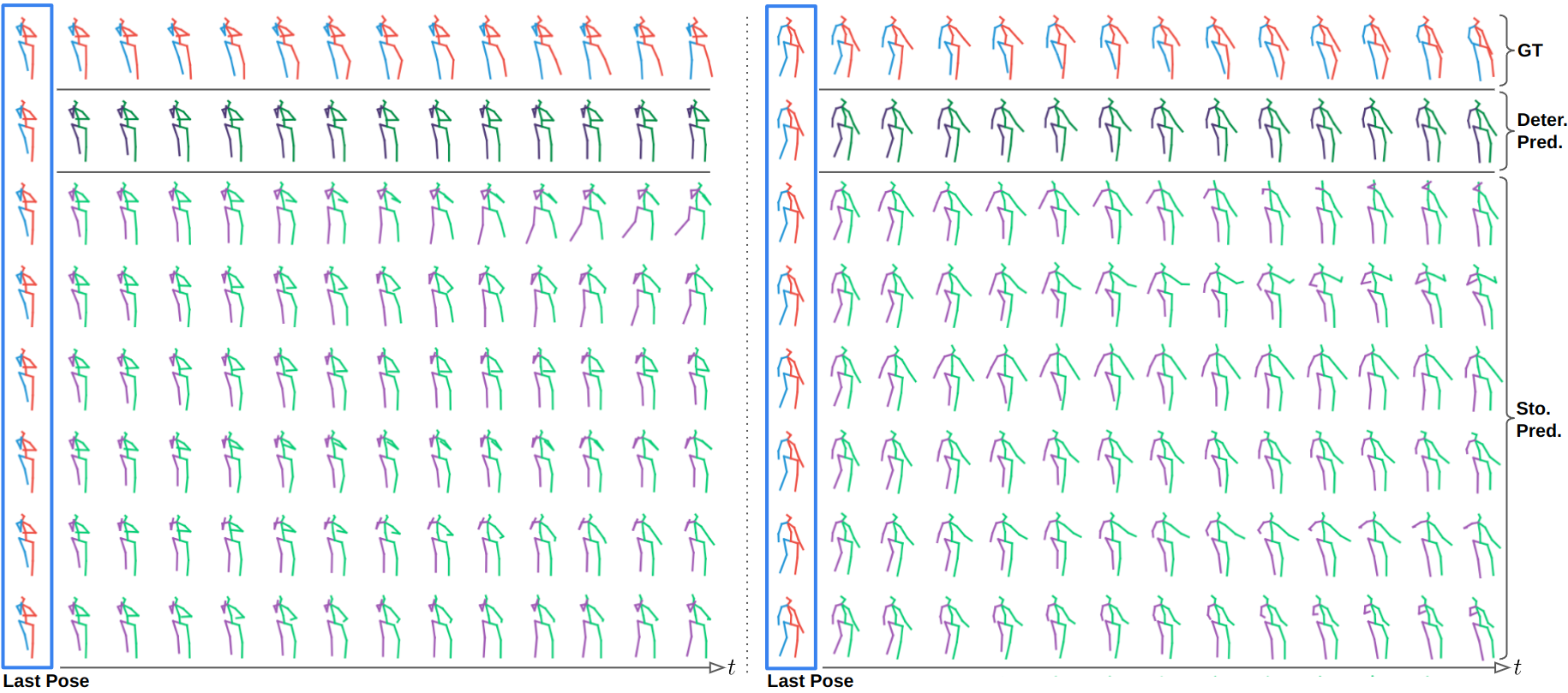}
    \caption{Deterministic (Deter.) and stochastic (Sto.) predictions from our transformer-based motion denoiser. Note that two results are given and divided based on the vertical dotted line. Predictions are obtained from observed motions labeled as `smoking' (left) and `walking' (right).}
    \label{fig:qual}
    \vspace{-3mm}
\end{figure*}

\subsection{Qualitative Results}
Figure~\ref{fig:qual} shows two example results from our transformer-based motion denoiser. Predictions on the left of the dotted line are obtained from the motion observation labeled as `smoking'. It is shown that the deterministic prediction is similar to the ground truth, while the stochastic predictions show the diversity between samples. But note that still the context of `smoking' looks remained in all samples. This phenomenon is also observed from the predictions on the right, which are obtained from the motion observation of `walking'. While its deterministic prediction resembles the ground truth, the stochastic predictions are diverse and contain the context of `walking'. 
For better visualization, please refer to our supplementary video.

\section{Conclusion} \label{sec:con}
In this work, we study the potential of diffusion probabilistic models for 3D human motion prediction tasks.
We propose two types of diffusion models based on the transformers, which understand the motion's spatial and temporal information in series or parallel.
Since the diffusion model is originally a generative model, its main usage would be for the stochastic motion prediction task. 
But once it is trained, we show that it can also be used in deterministic prediction if all randomness in its denoising process is ignored.

To show the effectiveness of diffusion models in both deterministic and stochastic motion prediction tasks, we conduct experiments based on various metrics. 
Results from deterministic prediction show that the diffusion model is not superior to the state-of-the-art. But it is shown that our long-term (1000ms) prediction performance is better than other transformer-based approaches. 
When it comes to evaluating stochastic predictions, it is conventional to suggest metrics measuring both likelihood and diversity.
However, we claim that the conventional metrics for measuring the likelihood do not represent how much the samples are distributed near the plausible motion, since they measure the \textit{minimum} distance between samples and ground truth.
Therefore, we suggest additional metrics to measure the mean and standard deviation of that distances, and the results show that our diffusion models can properly balance the trade-off between diversity and likelihood.

Although our results would provide nice answers to our first question -- can we use diffusion probabilistic models for 3D motion prediction? -- the most concerning disadvantage of a diffusion model is its sampling frequency. 
Since our diffusion model requires a $K=20$ number of denoising processes to obtain prediction samples, this might occur a bit high latency. To overcome this issue, one might consider recent works for efficient sampling \cite{salimans2021progressive}, which would be our future work, such that efficient 3D human motion prediction can be made for various real-time applications.

%%%%%%%%%%%%%%%%%%%%%%%%%%%%%%%%%%%%%%%%%%%%%%%%%%%%%%%%%%%%%%%%%%%%%%%%%%%%%%%%

%%%%%%%%%%%%%%%%%%%%%%%%%%%%%%%%%%%%%%%%%%%%%%%%%%%%%%%%%%%%%%%%%%%%%%%%%%%%%%%%

\section*{ACKNOWLEDGMENT}
This work was supported by the Institute of Information \& communications Technology Planning \& Evaluation (IITP) grant funded by the Korea government (MSIT) (No.2020-0-01336, Artificial Intelligence Graduate School Program (UNIST)), 
and funded by Marie Sklodowska-Curie Action Horizon 2020 (Grant agreement No. 955778) for project `Personalized Robotics as Service Oriented Applications' (PERSEO).
%%%%%%%%%%%%%%%%%%%%%%%%%%%%%%%%%%%%%%%%%%%%%%%%%%%%%%%%%%%%%%%%%%%%%%%%%%%%%%%%

\addtolength{\textheight}{-4cm}   % This command serves to balance the column lengths

%%%%%%%%%%%%%%%%%%%%%%%%%%%%%%%%%%%%%%%%%%%%%%%%%%%%%%%%%%%%%%%%%%%%%%%%%%%%%%%%

\bibliography{main}
\bibliographystyle{IEEEtran}

\end{document}